# Auditable Claims about AI Agents

Yue Zhao[1] (✉) [0000-0003-3401-4921], Jiate Li[1] [0009-0009-7829-0849], Li Li[1] [0009-0003-2007-2706], Yi Nian[1] [0000-0002-4387-5574], Jinbo Liu[2] [0009-0002-3224-2779], Xiaolin Zhou[2] [0009-0005-9548-3790], and Xiyang Hu[2] [0000-0001-7269-3828]

[1] University of Southern California, Los Angeles, CA, USA
{yue.z,jiateli,li.li02,yinian}@usc.edu
[2] Arizona State University, Tempe, AZ, USA
{jinboliu,xzhou226,xiyanghu}@asu.edu

**Abstract.** Organizations make claims about their AI agents: a person approves every external email, every action is logged, an evaluation shows the agent is safe to deploy. Article 12 of the EU AI Act requires high-risk systems to allow the automatic recording of events but does not say which records settle a given claim. The position is one sentence: to be checked, a claim about an agent must first name its policy, its scope, the records that would settle it, and who writes them. Adapting the preconditions of an assurance engagement, we call a claim auditable when these elements and a decision rule are fixed before any verdict and the records are obtainable. This extends the Policy Checkability dimension of our Auditable Agents framework from single actions to claims. Agents add three conditions: coverage by an independent record, authorization bound to each action's arguments, and completeness beyond integrity. Under an explicit model, we prove that support is impossible without each wherever its hypotheses hold. A claim-check table applies the method to six common claims, anchored in current NIST, IETF, and OWASP drafts. A worked case follows one claim through five evidence states. We close with a practice box and steps for operators, buyers, auditors, and standard setters.

## 1 Introduction

Claims about AI agents appear in system documentation, procurement answers, and incident reports, and others may later need to check them. Consider a firm whose document agent summarizes a matter folder and emails documents. The firm states that "a person approves every external email". On 6 March 2026, the agent sends email `tx-23`, with an attachment, to `press@example.net` after a confirmation in its chat interface. The mail gateway logs the delivery, but the log of the firm's authorization server holds no grant that covers it. Three questions become central: what exactly the claim covers, which records would settle it and who writes them, and what stays out of reach even when every record is present.

Rules about records are arriving. Article 12 of the EU AI Act requires high-risk AI systems to "technically allow for the automatic recording of events (logs) over the lifetime of the system" [9]. Its Article 14 requires that overseers be able to understand the system's limitations and stay aware of automation bias. These duties apply from 2 December 2027 for Annex III uses and from 2 August 2028 for Annex I products [10].

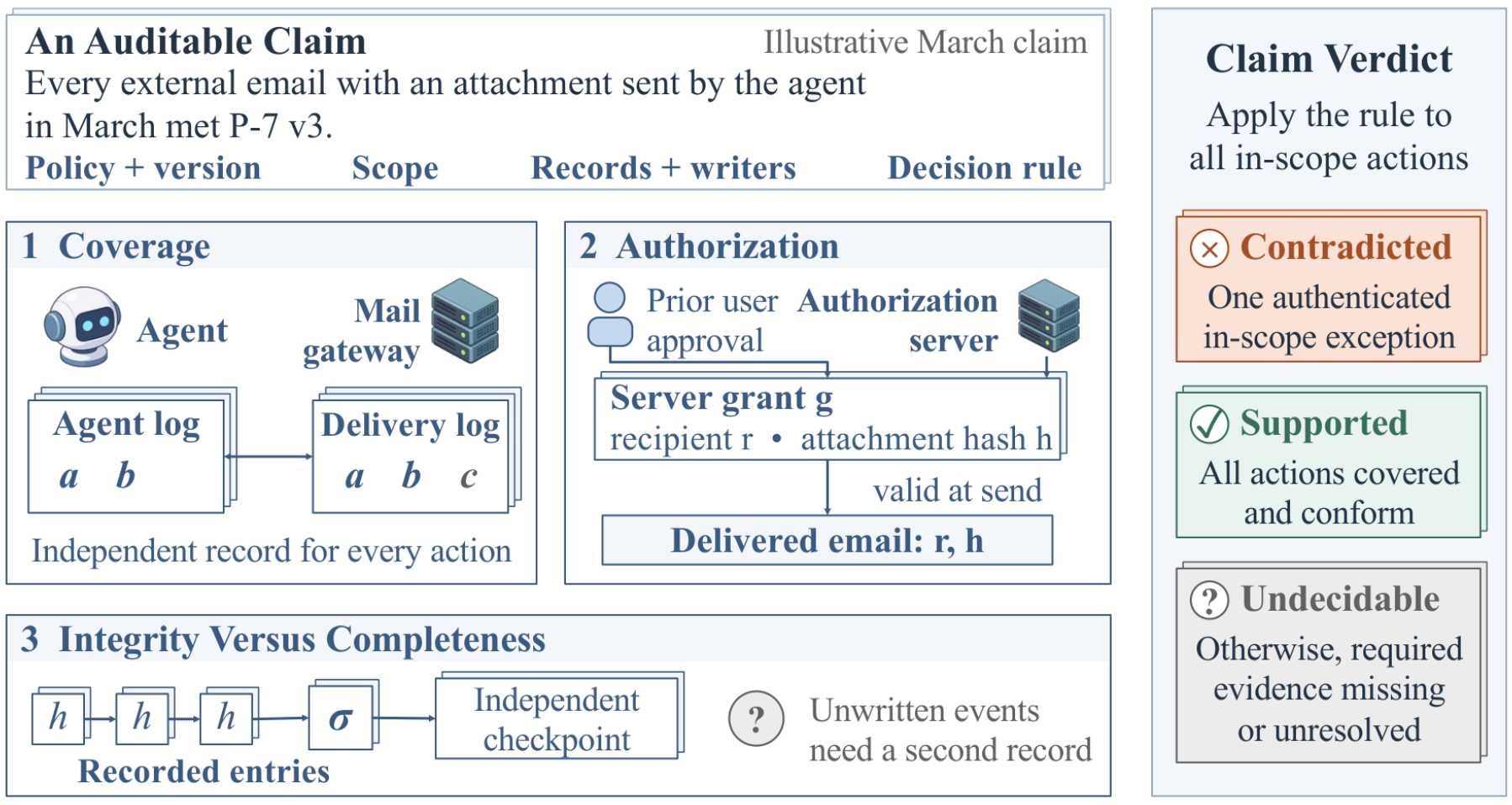

**Fig. 1.** From an auditable claim to an evidence-based verdict. Policy and version, scope, obtainable records with named writers, and a decision rule specify the claim before any verdict. Numbered panels mark three conditions agents add (Section 5): independent action coverage, prior authorization bound to the action's arguments, and completeness checked separately from integrity. Hash chains and signatures (σ) protect recorded entries. For the illustrative March claim, one authenticated in-scope exception (an email shown to violate P-7) contradicts it even if other records are missing. Support requires coverage and conformity for every in-scope email. Otherwise, missing or unresolved required evidence leaves the claim undecidable. These verdicts do not establish human intent or legal responsibility.

Standards bodies are drafting guidance on logging [13], documentation [1], evaluation [20], and agent authorization [14]. These documents say what a system should record or enable. Those we could read do not say, for a particular claim, which records would settle it or what stays out of reach.

Auditors already test claims this way: an assurance engagement checks a statement against suitable criteria, with evidence that must be obtainable [11]. We adapt that method to claims about agents. Agents change where the evidence comes from. Their actions may leave no record outside the agent's own log, the agent under control often records its own approvals, and signed logs are offered as audit trails.

*Position.* **To be checked, a claim about an agent must first name its policy, its scope, the records that would settle it, and who writes them.** The firm's statement names no policy version, period, record, or writer, so Section 4 rewrites it as an auditable claim. Even with every record present, the intent behind that confirmation stays out of reach.

This chapter continues our Auditable Agents framework [17] and moves its Policy Checkability dimension from single actions to claims. Written for auditors, engineers, and policy readers, it offers:

1. a working definition of an auditable claim (Section 2);
2. a claim-check table for six common claims, anchored in September 2026 drafts from NIST, the IETF, and OWASP (Section 3);
3. a worked case that moves one claim across five evidence states, with a filled claim-check worksheet (Section 4);
4. three conditions that agents add to established audit practice, with proofs, under an explicit model, of when support is impossible without each (Section 5);
5. steps for operators, buyers, auditors, and standard setters (Section 6), with a practice box on the companion repository and one evaluation claim.

Figure 1 summarizes the method, and Table 1 maps each part to the dimension of Auditable Agents it deepens.

**Table 1. Each part of the chapter deepens one dimension of Auditable Agents.** Dimensions as defined in [17]; Responsibility across chains of delegating agents is left to future work.

| Dimension | In Auditable Agents | Added in this chapter |
|---|---|---|
| Policy Checkability | Whether one action complies with a stated policy; a missing required field leaves it undecidable | A claim over many actions, fixed before any record is read (Definition 1), with three verdicts (Definition 2) |
| Action Recoverability | Action coverage and record fidelity | Coverage tested against a record the agent does not write (Condition 1) |
| Lifecycle Coverage | Approvals recorded as a distinct phase | A logged confirmation separated from a grant bound to the action's arguments (Condition 2) |
| Evidence Integrity | Records graded from none to signed | Integrity separated from completeness (Condition 3) |

The opening scenario and the worked case continue the external-email scenario of Auditable Agents. Two comments the first author submitted to NIST in September 2026 proposed documenting co-versioned tool schemas with the model [25] and naming the evaluated object in evaluation records [24]. The chapter proposes no standard. Its verdicts concern stated policies; legal compliance and responsibility remain separate judgments.

## 2 Auditable Claims

An audit here starts from a claim, such as "every email with an attachment that the agent sent outside the firm in March met policy P-7". Whether that claim can be checked depends first on the claim itself.

**Definition 1.** *A claim about an AI agent is* auditable *when four things are specified: policy and version, actions and period covered, required records obtainable from named writers, and the decision rule.*

A claim such as "our agent is safe" fails the definition before any record is read. Even a claim that meets it can be UNDECIDABLE when a required record is lost. Definition 2 gives the decision rule that C2 to C4 in Table 2 and the worked case share. C1, C5, and C6 keep the rules in that table.

**Definition 2.** *A claim that every in-scope action meets a stated policy is (i)* CONTRADICTED *when the required records establish that an authenticated in-scope action violated the policy. Where the violation is an absence, such as a missing grant, the record that should hold it must be complete for the relevant period. The claim is (ii)* SUPPORTED *when coverage evidence shows that every in-scope action was captured (Condition 1) and each one conforms. Otherwise it is (iii)* UNDECIDABLE*, and the audit names the missing record.*

Auditable Agents [17] judges one action: under a stated policy, its verdict is comply or violate. A missing required field can make that assessment undecidable, however many other events are logged. The chapter judges a claim over many actions, so support also needs coverage evidence.

Records differ in when they exist and in who writes them. CatchBench [26] distinguishes the declared configuration before a run (PRE), the steps recorded so far during the run (LIVE), and the record of the finished run (POST). A configuration can show what an agent could do; only a run record shows what it did. The writer may be the agent, an authorization server that issues grants, or a gateway that carries the action's effect. The definition names writers because whether a record can be trusted depends on who wrote it and who could change it.

*What we borrow.* The definition adapts the preconditions of an assurance engagement. Before accepting one, a practitioner under ISAE 3000 asks whether the criteria are suitable and whether the evidence needed for a conclusion can be obtained [11]. Our policy and version play the role of criteria, the actions and period the role of subject matter, and the records the role of evidence.

ISA 315 separates occurrence, that recorded events happened, from completeness, that every event was recorded [12]; the case in Section 4 tests both. We borrow this structure only. Our verdicts are not assurance opinions, and we leave aside materiality, sampling, and auditor independence, which a professional engagement would address.

*Related work.* Brundage et al. proposed institutional, software, and hardware mechanisms that make claims about AI development verifiable, audit trails included [2]. Ojewale et al. link tamper-evident audit trails for language model systems to governance records such as approvals, waivers, and attestations [18]. In a safety case, a developer argues that a system is safe [6]; an auditable claim is a statement that a third party checks against records.

Broader frameworks cover the internal audit process [21], governance, model, and application audits [16], and auditor access [3], and for agents, visibility [5]. We work one level below them and ask what a single claim about a deployed agent needs in order to be checked.

## 3 Six Common Claims and the Evidence They Need

Table 2 applies Definition 1 to six claims operators commonly make, with the passage in current guidance that bears on each. The anchors are a snapshot as of September 2026. They come from the OWASP Top 10 for Agentic Applications 2026 [19], the NIST AI 300-1 and AI 200-2 initial public drafts [1,20], and the IETF working-group draft draft-ietf-wimse-aims-00 [14].

We call the last AIMS; it is unrelated to the ISO/IEC 42001 management system that shares the abbreviation. A MUST in an Internet-Draft states a requirement for implementations of that draft; the draft is a work in progress and creates no legal duty. AIMS builds on OAuth, in which an *authorization server* issues each permission as a *grant*, carried in an access token that is signed or checked with the server (§10.3).

**Table 2. Six common claims, the evidence that settles each, and what stays out of reach.** Shaded rows carry the three conditions agents add (Section 5). Anchors quote drafts current in September 2026; the rules do not depend on them.

| Claim as stated | Decision rule and minimum evidence | Out of reach | Anchor |
|---|---|---|---|
| C1 “The deployed agent is the one we documented.” | Run-record identifiers of the model, co-versioned tool schemas, and each versioned system control (approval gate, execution policy) match those in force | That model documentation covers the system’s controls | NIST AI 300-1 §1: “only documentation of AI datasets and models”; systems “left to future work” |
| C2 “Every external action is logged.” | Each external action, with its effect-setting arguments and a correlation identifier, reconciled in both directions with an independent record that every relevant action must reach (Condition 1) | That no action bypassed every recorded path | AIMS §11: “reconstruct agent behavior and authorization context after execution”; OWASP ASI02; EU AI Act Art. 12(1) |
| C3 “A person approves every external transmission.” | A user-approved grant from the authorization server whose scope covers each such action and its arguments, valid when it ran (Condition 2) | What the approver understood; legal responsibility | AIMS §10.7: confirmations “do not by themselves constitute authorization”; OWASP ASI03 |
| C4 “The agent follows policy P.” | Policy text and version in force, evaluated on the C2 and C3 records; UNDECIDABLE if a required field is absent | Semantic rules, e.g., “discloses no identifying information” | AIMS §12: policy “versioned, reviewable”; §13: assessed on “recorded evidence” |
| C5 “Our logs are complete and tamper-proof.” | Signatures and chain checked against keys and a checkpoint held by another party; completeness tested against a record the agent does not write (Conditions 1 and 3) | That the log is tamper-proof; events no second record covers | AIMS §11: “Audit records MUST be tamper-evident” |
| C6 “Our evaluation shows the agent catches over-privilege.” | Evaluated object and tool version named; each data source compared with a trivial baseline; pinned rerun (practice box) | Transfer to a new deployment or data source | NIST AI 200-2 §4.3.1: benchmarks “often measure proxies contained in datasets” |

Two gaps recur across the rows. No field reveals an event that was never written (C2, C5). The seven minimum audit fields of AIMS §11 require the action requested without its effect-setting arguments, such as a recipient, which OWASP ASI02 asks to log. AIMS also leaves the policy model (§12) and compliance criteria (§13) to each deployment, which must therefore supply the versioned policy that C4 tests.

## 4 Worked Case: One Claim, Five Evidence States

This case continues the external-email scenario of Auditable Agents [17]; its records are constructed for teaching, and no agent was run.

*Setting.* A firm deploys a document agent that may read one matter folder, summarize it, and email documents. The firm uses the OAuth architecture AIMS describes [14], with the agent as an OAuth client. Its authorization server issues each grant as a signed token, one of the formats AIMS allows. The mail gateway computes the hash of every attachment it delivers (a fingerprint that changes if the file does). Policy P-7 requires each external email with an attachment to fall within a grant that a person approved before sending (Table 3).

*Making the claim auditable.* The operator states that "a person approves every external email". Table 3 rewrites it as an auditable claim: every email with an attachment that the agent sent to an external address in March met P-7, version 3. The claim is narrower than the statement. It leaves out emails without attachments, counts a prior batch approval, and covers only this version. Its three records come from three writers. Figure 2 shows two records the auditor receives.

**Table 3. The worksheet fixes the claim before any record is read; one exception then contradicts it.** The claim-check worksheet filled for the March claim. Mint rows are the four elements of Definition 1. The Card column names the question of the Auditability Card [17] whose answer supplies the field: Q1 actions recorded, Q2 phases such as approvals, Q3 checkable policies, Q4 responsibility chain, Q5 integrity, Q6 missing logs.

| Field | Entry | Card |
|---|---|---|
| Claim as stated | "A person approves every external email." | |
| Policy and version | P-7, version 3: each email with an attachment sent outside `example-firm.com` must fall within a grant from the authorization server that names its recipient and attachment hash. A signed-in person must approve the grant before sending (AIMS §10.4.1); grants an agent obtains on its own behalf do not count (§10.4.2). One grant may list several documents for one recipient, so a person approves a batch once. | |
| Actions and period | Emails with an attachment that the agent sent to external addresses, 1–31 March 2026 | |
| Records and writers | Agent log (the agent; supplied by the operator); grant log (the authorization server; exported by the identity team); delivery log (the mail gateway that every external email must pass; exported by IT), the independent record. The agent log is signed and hash-chained; no party outside the operator holds a checkpoint | Q1, Q2, Q4, Q5 |
| Decision rule | Definition 2 for P-7: the gateway log, if complete for March, is the coverage evidence; the server log should hold each grant | Q3 |
| Result | CONTRADICTED by `tx-23` (State C); `tx-31` and the fourth email of State E stay UNDECIDABLE | |
| Gaps and next steps | Log the correlation identifier on every send (State D); have the gateway record the sending client (State E); anchor the agent log with a checkpoint held by another party; check the grant at the gateway before delivery | Q6 |
| Out of reach | What `user:alvarez` understood or intended; who bears legal responsibility | |

```
{"corr": "tx-19",
 "agent": "doc-agent/2.4.1",
 "subject": "user:alvarez",
 "tool": "send_email",
 "args": {"to": "counsel@example.org",
   "attachment_sha256": "9f2c...e1"},
 "authz": {"grant": "g-5531",
   "issuer": "as.example-firm.com"},
 "t": "2026-03-04T10:02:14Z"}
```

```
{"corr": "tx-23",
 "agent": "doc-agent/2.4.1",
 "subject": "user:alvarez",
 "tool": "send_email",
 "args": {"to": "press@example.net",
   "attachment_sha256": "41ab...07"},
 "ui_confirmation": {"by": "user:alvarez",
   "t": "2026-03-06T15:40:02Z"},
 "t": "2026-03-06T15:40:05Z"}
```

**Fig. 2.** Two illustrative agent log records. Left: the agent names a grant that the server's record must confirm. Right: the agent logs only a confirmation in its own interface. Either way, the verdict turns on whether the server holds a prior user-approved grant covering the delivery.

*State A: configuration only (PRE).* The tool manifest (the list of tools the agent may call) grants `send_email` with no limit on the recipient domain and configures an approval step. This settles a configuration question: the agent may email any domain, although its task involves one outside counsel. The claim about March is UNDECIDABLE, because a configuration records no transmissions.

*State B: three records agree.* For email `tx-19`, the agent log names grant `g-5531`, and the authorization server logged at 10:02:11 that `user:alvarez` approved that grant for the same recipient and hash. The gateway logged delivery to that recipient at 10:02:15, with a hash it computed that matches. The token verifies and was valid at sending. This email met P-7, and the finding rests on three separately written records (C2, C3). The March claim still needs every other delivery matched.

*State C: confirmation without authorization.* Email `tx-23` carries a confirmation from the agent's chat interface and names no grant. The gateway logged its delivery to an external address, with a hash matching the agent's record. Complete for the period in which a grant could have been issued, the server's log holds none that covers it. This one authenticated exception makes the March claim CONTRADICTED, provided the server log is complete.

The record supports only a narrow statement: the interface logged a confirmation under the user's account. AIMS states that such confirmations "do not by themselves constitute authorization" and must be bound to a grant from the authorization server (§10.7). The requirement that each external email fall within an approved grant comes from P-7; AIMS leaves such policy to the deployment (§12). The verdict does not show that the user lacked intent to send. Auditable Agents [17] asks that approvals be recorded; the case adds the distinction between a local confirmation and a verifiable grant, which here decides the verdict.

*State D: a missing field.* For email `tx-31`, the agent log names no grant and lacks the correlation identifier, and the server's log for that hour was lost in a storage migration. The grant P-7 requires is absent from every surviving record, so the verdict for this email is UNDECIDABLE [17]. Because that absence is in the record and may not be in the world, reporting a P-7 violation would be an error. The lost log is a record-keeping finding in its own right.

*State E: a signed log and an extra email.* The supplied agent log is hash-chained (each entry includes a hash of the previous one) and signed, and it verifies. The gateway, however, shows four external emails with attachments from the agent's account, and the log records three. Verification authenticates the supplied entries; without an independent checkpoint, it cannot rule out a rewritten chain.

The agent may have sent the fourth email outside its logged path, the log may have dropped its entry, or a person may have used the same account. The server holds no grant matching that email, so its verdict stays UNDECIDABLE until another record, such as the gateway's client identifier, shows who sent it.

*What the case shows.* Table 4 summarizes the findings. State C alone contradicts the March claim, and each undecidable verdict names what the operator should record next time. Without State C, the claim would still lack support until every March delivery in the gateway log was matched to a grant.

**Table 4. One exception contradicts the March claim; three states stay undecidable.** The finding for each email and its effect on the March claim (Definition 2), shaded by finding. All records are illustrative.

| State | Evidence available | Finding | Reason |
|---|---|---|---|
| A (PRE) | Tool manifest and approval configuration | March claim UNDECIDABLE | A configuration records no transmissions |
| B, `tx-19` | Agent, server, and gateway logs agree on recipient and hash; server records the user's approval | This email meets P-7 | Three separately written records match |
| C, `tx-23` | Interface confirmation; gateway delivery; no grant in a complete server log | Violates P-7; March claim CONTRADICTED | A confirmation is not a grant; intent and responsibility stay open |
| D, `tx-31` | No grant reference or correlation identifier; server log lost | UNDECIDABLE for this email | The grant P-7 requires is absent from every surviving record |
| E, 4th email | Signed agent log verifies; gateway shows a fourth email with an attachment and no log entry or grant | UNDECIDABLE for this email | Sender unresolved: the account the gateway shows is shared, so it does not establish that the agent sent it |

## 5 Three Conditions Agents Add

The case turned on three tests that audit practice already applies. Agents change how each can be met, so we state each as a condition on the evidence.

*Model.* The *audited party*, the agent and the team that runs it, writes and may sign its own records. It may omit an in-scope action but never records one it did not take. An *independent writer*, such as the mail gateway or the authorization server run by other teams, is outside its control and records accurately what passes through it. The audited party cannot alter or forge such records. An approval such a writer records for a signed-in account is its holder's. A *complete* record holds every event of its kind in its period. A *world* is one history of the period, records included, that fits this model. A rule is *sound* if, from the records it reads alone, it returns SUPPORTED only when the claim of Definition 2 holds and CONTRADICTED only when it fails.

**Lemma 1 (Same records).** *A sound rule returns* UNDECIDABLE *in two worlds where it reads the same records if the claim holds in one but not the other.*
*Proof.* The rule returns one verdict for both. SUPPORTED would be wrong where the claim fails, and CONTRADICTED where it holds. □

In Fig. 3, the agent log hides the difference between the worlds; the independent gateway records the omitted send *c*.

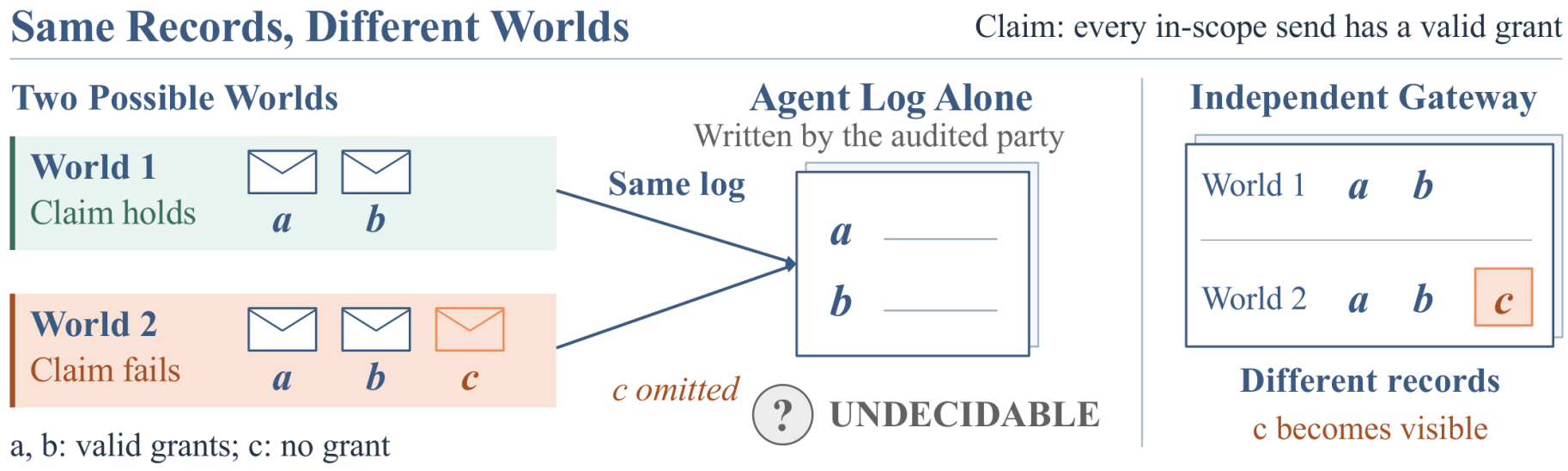

**Fig. 3. An omitted violation can leave the agent log unchanged.** For one agent and period, sends *a* and *b* have valid grants; *c* has none. The agent omits *c*, so a sound rule reading only its identical logs returns UNDECIDABLE in both worlds (Lemma 1). Here, an independent gateway records *c* as this agent's in-scope send, distinguishing the histories. Its entry establishes occurrence, leaving authorization unresolved. Together with a complete, independent grant log containing no grant covering *c*, this entry establishes CONTRADICTED (Proposition 1).

**Proposition 1 (One exception suffices).** *Let the policy require a grant for each in-scope action, as P-7 does. Suppose the records show an in-scope action, and the grant log, complete for the relevant period, holds no grant covering it. Then* CONTRADICTED *is correct in every world with these records.*
*Proof.* Independent writers are accurate and the audited party fabricates no action, so the action occurs in every such world. The complete grant log, which the audited party cannot alter, would hold any covering grant, so the claim fails. □

In State C, the gateway shows the delivery and the agent's own entry shows that it sent `tx-23`, so the proposition applies if the server log is complete.

*1. Coverage. Was every action recorded?* Support needs evidence that rules out an unrecorded violating action; a record of each action that the audited party does not write is one way to provide it. Auditors test the completeness assertion [12] by tracing from independent sources, such as confirmations from third parties, into the books. An agent's log has no independent counterpart by default: the agent writes it, and nothing else records each action. In State E, the agent log records three of the four emails the gateway shows.

That record can be a record of effects or the log of a gateway that every relevant action must pass (State B). Reconciling the agent log with it in both directions tests occurrence and completeness together. Condition 1 refines the action coverage that Auditable Agents measures under Action Recoverability.

**Corollary 1 (Coverage).** *Suppose every world could gain a violating in-scope action without changing the independent records a sound rule reads, as when it reads only the audited party's records. Then it never returns* SUPPORTED.
*Proof.* If the claim fails, SUPPORTED is wrong. Otherwise, add that action, which the audited party omits. The claim fails on the same records, so Lemma 1 applies. □

*2. Authorization. Did a person approve these arguments?* The evidence must establish the approval and the argument scope that the policy requires; P-7 requires a prior user-approved grant naming the recipient and attachment hash. A common control separates the person who approves a payment from the person who makes it, with the approval recorded independently. South et al. argue that agents need authenticated delegation, so that others can verify who authorized an agent to act [23]. Yet an agent often records its own approvals. In State C, the agent's interface wrote the confirmation, which records only an interface event attributed to an account. Neither a confirmation nor a grant shows what the person understood. For high-risk systems, Article 14 requires that overseers be able to understand the system's limitations and stay aware of automation bias [9]. A confirmation log is evidence of neither. Condition 2 refines Lifecycle Coverage, where Auditable Agents asks that approvals be recorded.

**Corollary 2 (Authorization).** *Let the policy require that a person approve each in-scope action with its effect-setting arguments. Suppose the independent records do not show such an approval for some action, as when only the audited party records it or they omit a value. Then no sound rule returns* SUPPORTED.
*Proof.* If the claim fails, SUPPORTED is wrong. Otherwise, take a twin world in which no person gave that approval. The independent records do not show it and can stay the same, as can the audited party's, so Lemma 1 applies. □

*3. Integrity versus completeness. Is a verified log complete?* Tamper-evident logs are often offered as audit trails; AIMS requires them (§11). A signature authenticates the entries supplied, and a hash chain anchored with an independent party exposes later edits to its entries [7,22]. In a recent preprint, a hash chain over the log that an agent harness (the software running the agent) keeps about itself flagged none of 28 injected omissions and fabrications. A log kept outside the harness flagged all 28 [8]. Condition 3 refines Evidence Integrity, which Auditable Agents grades from none through append-only and hash-chained to signed [17].

**Corollary 3 (Integrity versus completeness).** *Under the premise of Corollary 1, signatures, hash chains, and independent checkpoints of the entries the audited party submits cannot justify* SUPPORTED.
*Proof.* An omitted violating action changes neither the audited party's signed entries nor the checkpoint of those entries, and both still verify. The two worlds remain consistent with the model, so Corollary 1 applies. □

*Asymmetry.* Contradiction can rest on one authenticated action and a complete grant log (Proposition 1). Support needs independent records that would show any violating action (Corollary 1). The corollaries give necessary conditions only; sufficiency needs more, such as a gateway that no action can bypass.

*Limits of a verdict.* A verdict concerns a stated policy; it does not by itself decide legal compliance or liability.

*Alternative views.* We answer three objections. *A signed, tamper-evident log is enough*: it cannot show events it never recorded (Corollary 3). *A person should approve every action one at a time*: per-action approval may invite automation bias, and a grant for a scoped batch stays checkable (Condition 2). *Article 12 logging suffices*: it requires recording but names no records or writers that settle a claim (Definition 1).

**From Theory to Practice: Checking a Claim and Reading an Evaluation Score**

**Repository.** The chapter's companion repository, `https://github.com/yzhao062/auditable-claims`, holds the blank claim-check worksheet, the records of Fig. 2, and a pinned rerun of the tool below; Table 3 is the filled worksheet for the case. Code is MIT-licensed; text is CC BY 4.0.

**Tool.** CatchBench 0.1.2 [26] (MIT license, `https://github.com/yzhao062/catchbench`). Its PRE board scores methods that flag a declared agent configuration granting more privilege than its task needs. It covers 1,187 configurations from six sources and runs offline in about a second.

**Run.** The repository pins every dependency. In a fresh Python 3.12 environment, its script reproduced the full output, excerpted below, byte for byte.

```
git clone --branch v1.0 https://github.com/yzhao062/auditable-claims
cd auditable-claims/reproduce/catchbench-pre
./run.sh
```

**Output (excerpt, F1 by source).**

```
method              crewai   n8n   mcp injecagent sweagent synthetic overall
flag_all             0.388 0.154 0.654      0.750    0.574     0.763   0.601
owasp_asi_combined   0.448 0.411 0.644      0.961    0.570     0.842   0.654
label source: crewai, n8n, mcp (llm_judge); injecagent (roster_relabel);
              sweagent (declared_minus_used); synthetic (synthetic_inject)
```

**Reading.** F1 is a score from 0 to 1 that both misses and false alarms lower. `owasp_asi_combined` is the union of five rules that cite OWASP guidance. Its pooled F1 of 0.654 hides a range from 0.411 to 0.961. The rule that flags everything, `flag_all`, is a trivial baseline; its F1 varies from 0.154 to 0.763 because the share of over-privileged configurations differs by source. The pooled score mixes four label processes, and the tool itself says to read per source.

**What it supports.** The rule beats that baseline on four sources by 0.06 to 0.26 and falls just below it on mcp and sweagent. On injecagent, a rule that reads only declaration order scores 1.000, so there the score reflects how the corpus was built [26]. None of this shows that the rule will find over-privilege in a new deployment.

## 6 What Each Reader Can Do

Auditable Agents proposed adoption paths for papers, benchmarks, frameworks, and skill ecosystems [17]. Claims about deployed agents concern four other groups of readers.

*Operators and developers.* Write each public claim in the form of Definition 1 before publishing it, and keep the filled worksheet with it. Route consequential actions through a component the agent does not control, such as a gateway, and let that component write the counterpart record. Ask a person to approve a scoped batch of actions through the authorization server, rather than confirm each action in the agent's interface. Publishing the Auditability Card beside the claim tells readers which records exist.

*Buyers and procurement teams.* Ask vendors to state claims in that form and to name the writer of each record. Ask whether the buyer, or an auditor it appoints, can obtain those records. Read "tamper-proof logs" as claim C5 and ask which record the agent does not write. For an evaluation claim, ask for results per data source against a trivial baseline, with the evaluated version named (C6).

*Auditors.* Agree on the decision rule with the operator before examining records, as an assurance engagement requires [11]. Report UNDECIDABLE together with the record that would have settled the question, and report a lost log as a record-keeping finding apart from the verdict (State D).

*Standard setters.* Three additions to current guidance would make more claims decidable. The second and third qualify the book roadmap's advice to escalate high risks to human authorities and to log all interactions [15].

1. **Versioned system controls (C1).** NIST AI 300-1 leaves entire systems to future work, and its profiles may add fields but not change the base templates [1]. Approval gates and execution policies thus have no defined field. The 300-1 comment proposed asking what decides whether an action the model emits takes effect [25].
2. **Authorization bound to arguments (C3).** AIMS binds a local confirmation to a grant and notes that "Additional specification or design work may be needed" for interactions with the user during execution (§10.7). A profile could bind each approval to the exact arguments it covers, and the AIMS minimum audit fields could include them.
3. **A named second writer (C2, C5).** Guidance could ask each deployment to name a second writer, such as the authorization server or a gateway, for every class of consequential action, and to reconcile the two routinely. That record should keep only the fields the decision rule needs, under access and retention limits (AIMS §10.9). Accountability logs kept by smart contracts, which the roadmap names as one candidate for assigning responsibility [15], need the same second writer.

## 7 Summary, Limitations, and Discussion Questions

*Summary.* A claim about an AI agent is auditable when its policy, scope, records with their writers, and decision rule are fixed before the check, and the records are obtainable. Agents add three conditions: coverage, authorization bound to arguments, and completeness beyond integrity. Under the model and each corollary's hypotheses, support is impossible without them. The filled worksheet in Table 3 records one such check, and Section 6 turns the conditions into requests that operators, buyers, auditors, and standard setters can make.

*Limitations.* The anchors are drafts, and logging standards are still moving. We could not read ISO/IEC 24970, at the final-draft stage [13], or prEN 18229-1, under approval [4]. The Article 12 and 14 duties discussed here apply to high-risk AI systems, and whether an agent qualifies depends on its use or product. The chapter omits privacy and data minimization [17], responsibility across chains of delegating agents, and semantic policies that record schemas cannot decide.

Our case measures nothing, and the box covers only the offline PRE board of our own benchmark. We did not test the method. A test would seed known violations with complete reference records, then delete fields, records, or the independent record and re-sign a trimmed log (States D and E). It would score correct verdicts, correct UNDECIDABLE answers, false CONTRADICTED verdicts, and time per check.

*Discussion questions.*

1. **Claims.** Rewrite a public statement about an AI agent's approval feature as an auditable claim, using the worksheet of Table 3. Which of the four elements did the statement lack, and which evidence would change your verdict?
2. **Silence.** Who should bear the uncertainty when a record is silent, and should silence ever count as a violation?
3. **Confirmation.** In State C, the interface logged a confirmation but no grant exists. When, if ever, should that confirmation carry legal or organizational weight?
4. **Second writers.** Who in your setting could keep an independent record of an agent's actions, and what privacy obligations would that record create?
5. **Evaluation.** A benchmark reports one pooled score for a safety detector. What would you ask before using it in a procurement decision?

**Disclosure of Interests.** The authors have no competing interests to declare. The chapter discusses the authors' own benchmark and papers and the first author's NIST comments.